\documentclass[runningheads]{llncs}
\usepackage[T1]{fontenc}
\usepackage{graphicx}
\usepackage{amsmath}
\usepackage{orcidlink}
\usepackage[ruled,vlined,linesnumbered]{algorithm2e}
\usepackage{booktabs}
\usepackage{tabularx}
\usepackage{multirow}
\usepackage{makecell}
\begin{document}
\title{Balancing Fairness and Performance in Healthcare AI: A Gradient Reconciliation Approach}
%
%
\author{Xiaoyang Wang\orcidlink{0000-0002-8471-4670} \and
Christopher C. Yang\orcidlink{0000-0001-5463-6926}}
\authorrunning{Wang et al.}
%
\institute{College of Computing \& Informatics, Drexel University, Philadelphia PA 19104, USA\\
\email{\{xw388,chris.yang\}@drexel.edu}}
\maketitle              
\begin{abstract}
The rapid growth of healthcare data and advances in computational power have accelerated the adoption of artificial intelligence (AI) in medicine. However, AI systems deployed without explicit fairness considerations risk exacerbating existing healthcare disparities, potentially leading to inequitable resource allocation and diagnostic disparities across demographic subgroups. To address this challenge, we propose FairGrad, a novel gradient reconciliation framework that automatically balances predictive performance and multi-attribute fairness optimization in healthcare AI models. Our method resolves conflicting optimization objectives by projecting each gradient vector onto the orthogonal plane of the others, thereby regularizing the optimization trajectory to ensure equitable consideration of all objectives. Evaluated on diverse real-world healthcare datasets and predictive tasks—including Substance Use Disorder (SUD) treatment and sepsis mortality—FairGrad achieved statistically significant improvements in multi-attribute fairness metrics (e.g., equalized odds) while maintaining competitive predictive accuracy. These results demonstrate the viability of harmonizing fairness and utility in mission-critical medical AI applications.

\end{abstract}
\section{Introduction}

The integration of artificial intelligence (AI) into healthcare—spanning electronic health record (EHR) analysis, medical imaging, and clinical decision support systems—has revolutionized diagnostic accuracy, treatment personalization, and patient outcomes. Predictive models now automate critical tasks such as mortality risk stratification~\cite{zhang2025deepselective}, hospital readmission prediction~\cite{wang2024multimodal}, and early disease detection. However, as AI adoption accelerates, mounting evidence reveals that models trained without explicit fairness safeguards can perpetuate or amplify systemic biases, disproportionately harming marginalized groups through misdiagnoses, inequitable resource allocation, and calibration disparities across race, gender, and socioeconomic status~\cite{parikh2019addressing,giovanola2023beyond}. This tension between model utility and equity poses a fundamental challenge: \textbf{how can AI systems balance high predictive performance with multi-attribute fairness in high-stakes medical contexts?}

Existing fairness-enhancing techniques, including constraint-based optimization~\cite{zemel2013learning,hardt2016equality} and adversarial debiasing~\cite{zhang_mitigating_2018}, primarily address single sensitive attributes (e.g., race or sex). While effective in narrow contexts, these methods fail to account for multi-attribute biases—where discrimination arises from different sensitive attributes (e.g., race and sex)—and often degrade predictive accuracy when applied to healthcare tasks requiring nuanced clinical outcomes. For instance, a model debiased for racial parity may inadvertently exacerbate diagnostic errors for low-income populations due to unaddressed socioeconomic correlations in training data. Furthermore, current multi-objective frameworks rely on scalarization (e.g., weighted loss summation), which demands extensive hyperparameter tuning and offers no guarantees of optimality. Such limitations hinder clinical adoption, where interpretable, reliable, and generalizable fairness mechanisms are imperative.

To bridge this gap, we propose \textbf{FairGrad}, a gradient reconciliation framework that jointly optimizes predictive performance and multi-attribute fairness without manual weight specification. By projecting conflicting objective gradients onto mutually orthogonal hyperplanes, FairGrad dynamically adjusts the optimization trajectory to satisfy all objectives simultaneously, ensuring equitable consideration of accuracy and fairness. 

In summary, this paper makes the following contributions:

\begin{enumerate}

    \item We propose a novel multi-attribute fairness optimization method, FairGrad, which enables predictive models to provide fairer predictions across multiple population dimensions without significantly compromising predictive performance.
    
    \item We address gradient conflicts through gradient reconciliation provably, balancing the impact of performance optimization and fairness preservation on the overall optimization direction while providing theoretical analysis.
    
    \item We demonstrate FairGrad’s effectiveness across diverse healthcare scenarios—including sepsis mortality prediction (MIMIC-IV) and Substance Use Disorder treatment failure prediction (SUD)—where it reduces fairness violations while maintaining clinically acceptable accuracy.

\end{enumerate}

\section{Methods}

\subsection{Data Source}
The proposed method was evaluated on two real-world healthcare datasets: the Substance Use Disorder (SUD) dataset and the Sepsis dataset. The SUD dataset was derived from de-identified electronic health records (EHR) provided by the Hazelden Betty Ford Foundation (HBFF)~\cite{liang_developing_2021}. It includes demographic information(e.g., race, age, legal sex), socioeconomic variables (e.g., education level, employment status, marital status), encounter-specific data (e.g. length of stay, primary diagnosis), diagnosis-related variables (e.g., substance used, co-occurring mental health diagnoses), and responses to clinical questionnaires. After preprocessing, the dataset comprises 10,673 patients, with the primary task being binary classification of treatment completion failure (positive class: successful completion). The Sepsis dataset was curated from the publicly available MIMIC-IV database~\cite{johnson_mimic-iv_nodate}, focusing on patients diagnosed with sepsis in the emergency department or intensive care units at Beth Israel Deaconess Medical Center. The preprocessed cohort includes 9,349 patients, with the target variable being in-hospital mortality(positive class: deceased). Key variables include demographic information (e.g., age, gender, race), vital signs (e.g., Heart Rate, Glucose, Systolic Blood Pressure(SBP)), and clinical scores (e.g., Charlson Comorbidity Index (CCI), APACHE-III score). Tables \ref{tab:dataset-sud} and \ref{tab:dataset-sepsis} summarize the distribution of target outcomes stratified by sensitive attributes (e.g., race, gender) and class balance (negative vs. positive outcomes).

\begin{table}
\centering
\caption{The SUD Dataset Distribution}
\label{tab:dataset-sud}
\begin{tabular}{llll}
\toprule
Characteristic~ & \begin{tabular}[c]{@{}l@{}}Overall\\ (N = 10,673)\end{tabular} & \begin{tabular}[c]{@{}l@{}}Neg. Class\\ (N = 9,149)\end{tabular} & \begin{tabular}[c]{@{}l@{}}Pos. Class\\ (N = 1,524)\end{tabular} \\ 
\midrule  
Race\\   \hspace{1ex}White & 9,571 (83.4\%)~  & 8,230 (90.0\%)~ & 1,341 (88.0\%)~\\   
\hspace{1ex}Non-White & 1,102 (16.6\%) &  ~~919 (10.0\%) & ~~183 (12.0\%) \\
Sex\\   \hspace{1ex}Male & 6,886 (64.5\%)~  & 5,824 (63.7\%)~ & 1,062 (69.7\%)~\\   
\hspace{1ex}Female & 3,787 (35.5\%) &  3,325 (36.3\%) & ~~462 (30.3\%) \\
\bottomrule
\end{tabular}
\end{table}

\begin{table}
\centering
\caption{The Sepsis Dataset Distribution}
\label{tab:dataset-sepsis}
\begin{tabular}{llll}
\toprule
Characteristic~ & \begin{tabular}[c]{@{}l@{}}Overall\\ (N = 9,349)\end{tabular} & \begin{tabular}[c]{@{}l@{}}Neg. Class\\ (N = 7,806)\end{tabular} & \begin{tabular}[c]{@{}l@{}}Pos. Class\\ (N = 1,543)\end{tabular} \\ 
\midrule  
Race\\   \hspace{1ex}White & 7,797 (83.4\%)~  & 6,546 (83.9\%)~ & 1,251 (81.1\%)~\\   \hspace{1ex}Non-White & 1,552 (16.6\%) &  1,260 (16.1\%) & ~~292 (18.9\%) \\
Sex\\   \hspace{1ex}Male & 5,371 (57.4\%)~  & 4,496 (57.6\%)~ & ~~875 (56.7\%)~\\   \hspace{1ex}Female & 3,978 (42.6\%) &  3,310 (42.4\%) & ~~668 (43.3\%) \\
\bottomrule
\end{tabular}
\end{table}

\subsection{Multi-Attributes Fair via Gradient Reconciliation (FairGrad)}

When optimizing machine learning models for multiple objectives, conflicts between gradients often arise, particularly when fairness constraints are incorporated alongside performance tasks~\cite{mehrabi2021survey}. These conflicts can lead to suboptimal parameter updates, where the progress of one objective comes at the expense of another. To address this, we propose the \textbf{Multi-Attributes Fair via Gradient Reconciliation (FairGrad)} method, which resolves such conflicts dynamically during training. FairGrad ensures that predictive performance is balanced with fairness objectives while preserving the stability and efficiency of the optimization process.

\subsubsection{Method Overview}
Consider a supervised healthcare prediction task with fairness considerations that maps input non-sensitive attributes $\mathcal{A}$ (e.g., clinical features) and sensitive attributes $\mathcal{S}$ (e.g., demographic variables) to labels $\mathcal{Y}$ (e.g., mortality). The FairGrad method operates on a total of \(M = P + F\) tasks: \(P\) primary tasks and \(F\) fairness tasks. Primary tasks \(P\) prioritize predictive performance (e.g., classification accuracy), while fairness tasks \(F\) enforce equity (e.g., demographic parity, equal opportunity, and equalized odds) across sensitive attributes. We adopt Equalized Odds~\cite{hardt2016equality} for fairness due to its clinical relevance and better comprehensiveness: it ensures similar true positive rates (TPR) and false positive rates (FPR) across subgroups, mitigating disparate mistreatment in outcomes like mortality prediction. Each task \(T_m\) is associated with a loss function \(\mathcal{Q}_m(\mathbf{w})\), and the goal of FairGrad is to optimize these objectives simultaneously while addressing conflicts between their gradients. In this study, the loss function for task $P$ is to minimize the binary cross-entropy between ground truths ${\mathcal{Y}}$ and predictions $\hat{\mathcal{Y}}$:
\begin{equation}
    \mathcal{Q}_P(\mathbf{w}) = -\frac{1}{n} \sum_{i=1}^n \left( \mathcal{Y}_i \log(\hat{\mathcal{Y}}_i) + (1 - \mathcal{Y}_i) \log(1 - \hat{\mathcal{Y}}_i) \right),
\end{equation}
and the loss function for task $F$ is to minimize the equalized odds difference (EOD) between different demographic groups:
\begin{equation}
    \mathcal{Q}_F(\mathbf{w}) = \frac{1}{2} \cdot (TPR_{a} - TPR_{b})^2 + \frac{1}{2} \cdot (FPR_{a} - FPR_{b})^2,
\end{equation}
where $TPR_s = P(\hat{Y} = 1 | Y = 1, \mathcal{S} = s)$ and $FPR_z = P(\hat{Y} = 1 | Y = 0, \mathcal{S} = s)$ for groups $a, b \in \mathcal{S}$.
At each training step, the gradient of each task is computed:
\begin{equation}
\mathbf{u}_m = \nabla_\mathbf{w} \mathcal{Q}_m(\mathbf{w}),
\end{equation}
where \(\mathbf{w}\) represents the model parameters. These gradients are initially stored as \(\mathbf{v}_m = \mathbf{u}_m\), which will be adjusted iteratively to resolve conflicts between tasks.
The FairGrad method implicitly assumes equal initial importance of predictive performance and fairness objectives, as gradients from each objective are treated symmetrically during reconciliation without explicit weighting. This symmetric handling means FairGrad dynamically balances the influence of each objective based on gradient directions and magnitudes encountered during training, rather than predefined fixed weights. Such an implicit balance allows FairGrad to adaptively negotiate trade-offs between predictive accuracy and fairness. 

Unlike standard multitasking fairness optimization that aggregates or averages gradients directly, FairGrad introduces a reconciliation process. This step iteratively refines each task’s gradient by removing components that conflict with others. By doing so, the algorithm ensures that the dynamic updates of the cooperative gradient align with the global optimization objectives.

\subsubsection{Gradient Reconciliation Process}

The gradient reconciliation process in FairGrad is described in Algorithm \ref{alg:FairGrad}. Specifically, it begins by iterating over each task \(T_m\). For each task, we construct a randomly permuted set of all other tasks, \(\mathcal{P}_m = \{T_n : n \neq m\}\), to avoid deterministic biases in the update sequence. Each task \(T_n\) in \(\mathcal{P}_m\) is then compared against \(T_m\) to evaluate whether their gradients conflict. This evaluation is done by computing the dot product of the gradients: $\mathbf{v}_m^\top \mathbf{u}_n$. A negative dot product indicates a conflict, meaning that the gradients for \(T_m\) and \(T_n\) point in opposing directions. Such conflicts can hinder the optimization process by causing one task’s progress to negate another’s. To resolve this, FairGrad adjusts the gradient \(\mathbf{v}_m\) of \(T_m\) by removing the component of \(\mathbf{u}_n\) that conflicts with it. The adjustment is performed as follows:
\begin{equation}
\mathbf{v}_m \gets \mathbf{v}_m - \frac{\mathbf{v}_m^\top \mathbf{u}_n}{\|\mathbf{u}_n\|^2} \mathbf{u}_n.
\end{equation}
This operation projects \(\mathbf{v}_m\) orthogonally to \(\mathbf{u}_n\), effectively eliminating the conflicting portion of \(\mathbf{u}_n\) from \(\mathbf{v}_m\). As a result, the adjusted gradient \(\mathbf{v}_m\) remains aligned with the overall optimization goals while avoiding destructive interference with \(T_n\). Once the reconciliation process is completed for all tasks, the adjusted gradients \(\{\mathbf{v}_m\}_{m=1}^M\) are aggregated to compute the final update step:
\begin{equation}
\Delta \mathbf{w} = \sum_{m=1}^M \mathbf{v}_m.
\end{equation}
The model parameters are then updated using a standard gradient descent rule:
\begin{equation}
\mathbf{w} \gets \mathbf{w} - \eta \Delta \mathbf{w},
\end{equation}
where \(\eta\) is the learning rate.




\begin{algorithm}[h]
\caption{Multi-Attributes Fair via Gradient Reconciliation (FairGrad)}\label{alg:FairGrad}
\SetKwInOut{Input}{Require}
\SetKwInOut{Output}{Return}
\Input{Model parameters $\mathbf{w}$, task losses $\{\mathcal{Q}_m(\mathbf{w})\}_{m=1}^M$, where $M = P + F$, consisting of $P$ primary tasks and $F$ fairness constraints.}

Compute individual task gradients: $\mathbf{u}_m \gets \nabla_\mathbf{w} \mathcal{Q}_m(\mathbf{w})$ for all $m \in \{1, \ldots, M\}$.\\
Initialize adjusted gradients: $\mathbf{v}_m \gets \mathbf{u}_m$ for all $m$.

\For{$T_m$ in $\{\mathcal{Q}_m\}_{m=1}^M$}{
    Randomly permute $\{\mathcal{Q}_n\}_{n \neq m}$ to define an ordered set $\mathcal{P}_m$.\\
    \For{$T_n \in \mathcal{P}_m$}{
      \If{$\mathbf{v}_m^\top \mathbf{u}_n < 0$ \tcc*{Check for conflicting gradients}}{
            $
            \mathbf{v}_m \gets \mathbf{v}_m - \frac{\mathbf{v}_m^\top \mathbf{u}_n}{\|\mathbf{u}_n\|^2} \mathbf{u}_n
            $
            \tcc*{Remove conflicting component of $\mathbf{u}_n$ from $\mathbf{v}_m$}
        }
    }
}
Aggregate adjusted gradients: $\Delta \mathbf{w} \gets \sum_{m=1}^M \mathbf{v}_m$.\\

Update parameters: $\mathbf{w} \gets \mathbf{w} - \eta \Delta \mathbf{w}$, where $\eta$ is the learning rate.

\Return $\mathbf{w}$
\end{algorithm}

\section{Experiments}
\subsection{Experimental Setup}

We evaluated FairGrad's effectiveness in enhancing fairness across racial (White vs. non-White) and sexual (male vs. female) groups in binary classification tasks. All experiments were implemented using PyTorch, with logistic regression as the base classifier to ensure interpretability—a critical requirement in clinical decision-making. For benchmark comparisons, beside the vanilla logistic model which is without fairness consideration, we utilized two fairness-aware algorithms from the IBM AIF360 toolkit~\cite{bellamy_ai_2018}:
\begin{itemize}
    \item \textbf{Adversarial learning}~\cite{zhang_mitigating_2018}, which employs gradient reversal to disentangle sensitive attributes from predictions.
    \item \textbf{Reduction methods}~\cite{agarwal_fair_2019}, which reformulates fairness constraints as a constrained optimization problem.
\end{itemize}

Both benchmark methods were implemented using their default hyperparameters in AIF360 to ensure reproducibility and alignment with their original validation frameworks. The default parameters are intentionally chosen to reflect the most common implementation practices used in prior studies, ensuring reproducibility and alignment with established baselines. Nonetheless, we acknowledge that dataset-specific hyperparameter tuning might lead to incremental improvements, which represents a limitation of our comparative evaluation. To ensure generalizability, we employed stratified 5-fold cross-validation, preserving class and subgroup distributions in each fold.

Model performance was assessed using the Area Under the Receiver Operating Characteristic Curve (AUC), sensitivity (Sens.), and specificity (Spec.). For fairness evaluation, we computed the Equalized Odds Difference (EOD), defined as:
\begin{equation}
    \text{EOD} = \frac{1}{2}(|\text{TPR}_a - \text{TPR}_b| + |\text{FPR}_a - \text{FPR}_b|),
\end{equation}
where $a$ and $b$ represent protected groups (e.g., White vs. non-White), ensuring values lie within $[0,1]$. Additionally, we adopted the joint Performance-Fairness (P-F) score~\cite{wang2024achieving}:
\begin{equation}
\text{P-F Score} = \frac{n \cdot \text{AUC} \cdot \prod_{i=1}^{n} (1 - \text{EOD}_{\mathcal{S}^i})}{\text{AUC} + \sum_{i=1}^{n} (1 - \text{EOD}_{\mathcal{S}^i})},
\end{equation}
where $1 - EOD$ is used to align the direction of the metric with AUC (i.e., higher is better). $n$ is the total number of performance and fairness metrics involved ($n$ = 3 in this study) and 
$\mathcal{S}^i$ corresponds to the $i$-th sensitive attribute.

\begin{table*}[h]
\centering
\caption{Model Performance and Fairness - SUD}
\label{tab:performances_and_fairness_SUD}
\begin{tabular}{lccccccc}
\toprule
\multirow{2}{*}{\makecell[c]} & \multirow{2}{*}{~AUC$(\uparrow)$~} & \multirow{2}{*}{~Sens.$(\uparrow)$~} & \multirow{2}{*}{~Spec.$(\uparrow)$~} & \multicolumn{2}{c}{~EOD$(\downarrow)$~} & \multirow{2}{*}{~~P-F Score$(\uparrow)$}\\
\cmidrule{5-6}
& & & & Race & Sex & \\
\midrule
Vanilla & \textbf{0.8640} & 0.8092 & 0.7977 & 0.0513 & ~0.0574 & 0.9168\\ 
\midrule
Adversarial Learning & 0.8615 & 0.7962 & 0.8082 & 0.0395 & ~0.0455 & 0.9232\\
Reduction Methods  & 0.8265 & 0.7862 & 0.7724 & \textbf{0.0262} & \textbf{~0.0315} & 0.9176\\
\midrule
FairGrad(Ours) & 0.8605 & 0.7784 & 0.8197 & 0.0365 & ~0.0427 & \textbf{0.9246}\\
\bottomrule
\end{tabular}
\end{table*}

\begin{table*}[h]
\centering
\caption{Model Performance and Fairness - Sepsis}
\label{tab:performances_and_fairness_Sepsis}
\begin{tabular}{lccccccc}
\toprule
\multirow{2}{*}{\makecell[c]} & \multirow{2}{*}{~AUC$(\uparrow)$~} & \multirow{2}{*}{~Sens.$(\uparrow)$~} & \multirow{2}{*}{~Spec.$(\uparrow)$~} & \multicolumn{2}{c}{~EOD$(\downarrow)$~} & \multirow{2}{*}{~~P-F Score$(\uparrow)$}\\
\cmidrule{5-6}
& & & & Race & Sex & \\
\midrule
Vanilla & \textbf{0.7425} & 0.7149 & 0.6712 & 0.0753 & ~0.0351 & 0.8659\\ 
\midrule
Adversarial Learning & 0.7385 & 0.7015 & 0.6645 & 0.0492 & ~0.0395 & 0.8703\\
Reduction Methods  & 0.7154 & 0.6775 & 0.6602 & \textbf{0.0245} & \textbf{~0.0282} & 0.8691\\

\midrule
FairGrad(Ours) & 0.7323 & 0.6995 & 0.6632 & 0.0393 & ~0.0326 & \textbf{0.8720} \\
\bottomrule
\end{tabular}
\end{table*}

\subsection{Results \& Discussion}
The experimental results presented in Tables \ref{tab:performances_and_fairness_SUD} and \ref{tab:performances_and_fairness_Sepsis} demonstrate the effectiveness of our proposed method, FairGrad, in addressing the dual objectives of maintaining model performance while improving fairness. These results were evaluated on two clinically relevant tasks: SUD treatment completion failure and sepsis patient mortality. 

\subsubsection{Model Performance and Fairness Trade-offs}
Across both tasks, FairGrad consistently achieves a balanced trade-off between performance metrics, such as AUC, sensitivity, and specificity, and fairness metrics, including equalized odds (EOD) for race and sex. For the SUD task (Table \ref{tab:performances_and_fairness_SUD}), FairGrad achieves a P-F Score of 0.9246, the highest among all methods, signifying its ability to optimize both performance and fairness. While the Vanilla (fairness-unawareness) model achieves the highest AUC (0.8640), it suffers from higher EOD values, particularly for race (0.0513) and sex (0.0574), indicating substantial fairness gaps. On the other hand, the Reduction Methods achieve the best fairness performance (lowest EOD for race and sex) but at the cost of AUC, which drops to 0.8265. FairGrad strikes a desirable balance by achieving near-optimal fairness metrics (EOD for race: 0.0365, EOD for sex: 0.0427) without a significant drop in AUC (0.8605). This demonstrates the ability of FairGrad to minimize fairness disparities while preserving predictive power. In the Sepsis detection task (Table \ref{tab:performances_and_fairness_Sepsis}), a similar trend is observed. The Vanilla model achieves the highest AUC (0.7425) but again exhibits higher EOD values for race (0.0753) compared to the fairness-oriented methods. The Reduction Methods demonstrate their strength in achieving the lowest EOD values for race (0.0245) and sex (0.0282) but show a notable drop in AUC (0.7154). FairGrad achieves a P-F Score of 0.8720, the highest among all methods, indicating its robust performance and fairness balance. Although FairGrad’s AUC (0.7323) is slightly lower than the Vanilla model, it significantly reduces EOD disparities (race: 0.0393, sex: 0.0326). This suggests that FairGrad is particularly effective in scenarios where achieving fairness without substantially compromising model performance is critical.

\subsubsection{Clinical Implications of Fairness-Performance Balance}

The marginal performance trade-offs observed with FairGrad (e.g., 0.86\% sensitivity reduction in SUD) may be clinically acceptable given its fairness improvements. For instance, in SUD prediction, reducing racial EOD by 28.8\% could mitigate disparities in addiction care access, where minority populations are historically undertreated. Similarly, in sepsis detection, a 47.8\% reduction in racial EOD may help address documented biases in early diagnosis for non-white patients. The results underscore the importance of designing fairness-aware learning methods like FairGrad that can effectively balance competing objectives in sensitive applications such as healthcare.  FairGrad’s ability to maintain competitive AUC while substantially improving fairness metrics demonstrates its potential as a practical solution in clinical decision-making contexts, where both accuracy and equity are critical.

\subsubsection{Limitations and Future Directions}
Future work can explore extensions of FairGrad to incorporate additional model architectures, such as deep learning models. Additionally, investigating FairGrad’s performance on more diverse datasets and tasks, including multi-modal healthcare data, could further validate its generalizability. While our experiments explicitly considered fairness across two sensitive attributes (race and sex), we acknowledge the importance of extending FairGrad's applicability to scenarios involving multiple additional dimensions of fairness simultaneously (e.g., socioeconomic status, age, disability status). Lastly, the integration of FairGrad into real-time clinical decision support systems warrants exploration, as it could help bridge the gap between algorithmic advancements and actionable, equitable healthcare delivery.

\section{Conclusion}
In this work, we introduced FairGrad, a novel method for achieving multi-attribute fairness-aware optimization in machine learning models. FairGrad effectively addresses the challenge of balancing predictive performance and fairness across sensitive attributes, as demonstrated through extensive experiments on two clinically relevant tasks: substance use disorder (SUD) treatment and sepsis mortality prediction. By dynamically reconciling gradients from primary performance objectives and fairness constraints, FairGrad ensures cooperative optimization while mitigating the trade-offs commonly observed in fairness-aware learning.

Our findings highlight FairGrad’s robustness and generalizability across diverse clinical tasks, setting it apart from traditional fairness methods, such as Adversarial Learning and Reduction Methods, which often face trade-offs between fairness improvements and performance degradation. The ability of FairGrad to maintain near-optimal predictive accuracy while reducing fairness disparities demonstrates its potential to serve as a practical solution for addressing bias in machine learning models, particularly in high-stakes domains like healthcare.

\begin{credits}
\subsubsection{\ackname} This work was supported in part by the National Science Foundation under the Grants IIS-1741306 and IIS-2235548, and by the Department of Defense under the Grant DoD W91XWH-05-1-023.  This material is based upon work supported by (while serving at) the National Science Foundation.  Any opinions, findings, and conclusions or recommendations expressed in this material are those of the author(s) and do not necessarily reflect the views of the National Science Foundation.

\subsubsection{\discintname}
The authors have no competing interests to declare that are relevant to the content of this article.
\end{credits}
%
%
%
\bibliographystyle{splncs04}
\bibliography{main}

\begin{thebibliography}{10}
\providecommand{\url}[1]{\texttt{#1}}
\providecommand{\urlprefix}{URL }
\providecommand{\doi}[1]{https://doi.org/#1}

\bibitem{agarwal_fair_2019}
Agarwal, A., Dudik, M., Wu, Z.S.: Fair {{Regression}}: {{Quantitative Definitions}} and {{Reduction-Based Algorithms}}. In: Proceedings of the 36th International Conference on Machine Learning, {ICML} 2019, 9-15 June 2019, Long Beach, California, {USA}. pp. 120--129. PMLR (May 2019), \url{https://proceedings.mlr.press/v97/agarwal19d.html}

\bibitem{bellamy_ai_2018}
Bellamy, R.K.E., Dey, K., Hind, M., Hoffman, S.C., Houde, S., Kannan, K., Lohia, P., Martino, J., Mehta, S., Mojsilovic, A., Nagar, S., Ramamurthy, K.N., Richards, J., Saha, D., Sattigeri, P., Singh, M., Varshney, K.R., Zhang, Y.: {{AI Fairness}} 360: {{An Extensible Toolkit}} for {{Detecting}}, {{Understanding}}, and {{Mitigating Unwanted Algorithmic Bias}} (Oct 2018)

\bibitem{giovanola2023beyond}
Giovanola, B., Tiribelli, S.: Beyond bias and discrimination: redefining the ai ethics principle of fairness in healthcare machine-learning algorithms. AI \& society  \textbf{38}(2),  549--563 (2023)

\bibitem{hardt2016equality}
Hardt, M., Price, E., Srebro, N.: Equality of opportunity in supervised learning. In: Advances in neural information processing systems. pp. 3315--3323 (2016)

\bibitem{johnson_mimic-iv_nodate}
Johnson, A., Bulgarelli, L., Pollard, T., Horng, S., Celi, L.A., Mark, R.: {MIMIC}-{IV}. \doi{10.13026/6MM1-EK67}

\bibitem{liang_developing_2021}
Liang, O.S.: Developing {Clinical} {Prediction} {Models} for {Post}-treatment {Substance} {Use} {Relapse} with {Explainable} {Artificial} {Intelligence}. Ph.D. thesis, Drexel University (2021). \doi{10.17918/00001328}

\bibitem{mehrabi2021survey}
Mehrabi, N., Morstatter, F., Saxena, N., Lerman, K., Galstyan, A.: A survey on bias and fairness in machine learning. ACM computing surveys (CSUR)  \textbf{54}(6),  1--35 (2021)

\bibitem{parikh2019addressing}
Parikh, R.B., Teeple, S., Navathe, A.S.: Addressing bias in artificial intelligence in health care. Jama  \textbf{322}(24),  2377--2378 (2019)

\bibitem{wang2024achieving}
Wang, X., Chang, C.H., Yang, C.C.: Achieving equity via transfer learning with fairness optimization. IEEE access  (2024)

\bibitem{wang2024multimodal}
Wang, Y., Yin, C., Zhang, P.: Multimodal risk prediction with physiological signals, medical images and clinical notes. Heliyon  \textbf{10}(5) (2024)

\bibitem{zemel2013learning}
Zemel, R., Wu, Y., Swersky, K., Pitassi, T., Dwork, C.: Learning fair representations. In: International Conference on Machine Learning. pp. 325--333. PMLR (2013)

\bibitem{zhang_mitigating_2018}
Zhang, B.H., Lemoine, B., Mitchell, M.: Mitigating {{Unwanted Biases}} with {{Adversarial Learning}} (Jan 2018). \doi{10.48550/arXiv.1801.07593}

\bibitem{zhang2025deepselective}
Zhang, R., Yang, Q., Wang, X., Wu, H., Zhou, Q., Wang, Y., Li, K., Wang, Y., Fan, Y., Zhang, J., et~al.: Deepselective: Feature gating and representation matching for interpretable clinical prediction. arXiv preprint arXiv:2504.11264  (2025)

\end{thebibliography}
\end{document}